\documentclass{article}
\usepackage{ijcai20}  

\usepackage{times}  
\usepackage{helvet} 
\usepackage{courier}  
\usepackage[hyphens]{url}  
\usepackage{graphicx} 
\usepackage{graphicx}  
\usepackage{framed}
\usepackage{xspace}
\usepackage{xcolor}
\usepackage{todonotes}

\usepackage{amsmath,amsfonts,amsthm,bm} 

\usepackage{graphicx}
\usepackage{lipsum}
\allowdisplaybreaks
\usepackage{cleveref}
\usepackage{lscape}
\usepackage{pgfgantt}

\usepackage{tabularx}
\usepackage{threeparttable}  
\usepackage{tikz}
\usepackage{pgfplots}
\usepackage{multirow}
\usepackage{algorithmicx}
\usepackage{algorithm,algpseudocode}
\algnewcommand\algorithmicforeach{\textbf{for each}}
\algdef{S}[FOR]{ForEach}[1]{\algorithmicforeach\ #1\ \algorithmicdo}

\usepackage{amssymb}
\definecolor{azure}{rgb}{0.0, 0.5, 1.0}
\definecolor{blau_2b}{RGB}{0,131,204}

\usepackage{enumitem}

\usepackage{soul}
\usepackage{breakcites}
\usepackage{empheq}

\newcommand{\vw}{\boldsymbol{w}}

\newcommand{\vx}{\boldsymbol{x}}
\newcommand{\vy}{\boldsymbol{y}}

\newcommand{\vs}{\boldsymbol{s}}

\newcommand{\vb}{\boldsymbol{b}}

\newcommand{\vTheta}{\boldsymbol{\Theta}}
\newcommand{\vphi}{\boldsymbol{\phi}}

\newcommand{\poorya}[1]{{\color{black} #1}}
\newcommand{\reza}[1]{{\color{black} #1}}

\newcommand{\tb}[1]{\textbf{#1}}

\newcommand{\mdag}{\textsc{Daager}\ }

\newcommand{\va}{\pmb{a}}

\title{Learning to Multi-Task Learn for Better Neural Machine Translation}

\author{
Poorya Zaremoodi
\and
Gholamreza Haffari
\affiliations
Faculty of Information Technology \\
Monash University,  Australia\\
\emails
firstname.lastname@monash.edu
}

\begin{document}

\maketitle

\begin{abstract}
Scarcity of parallel sentence pairs is a major challenge for training high quality neural machine translation (NMT) models in bilingually low-resource scenarios, as NMT is data-hungry.
Multi-task learning is an elegant approach to inject linguistic-related inductive biases into NMT, using auxiliary syntactic and semantic tasks, to improve generalisation. 
The challenge, however, is to devise effective \emph{training schedules}, prescribing when to make use of the auxiliary tasks during the training process to fill the knowledge gaps of the main translation task, a setting referred to as \emph{biased-MTL} .
%
%
Current approaches for the training schedule are based on hand-engineering heuristics, 
whose effectiveness vary in different MTL settings.
We propose a novel framework for \emph{learning} the training schedule, ie \emph{learning to multi-task learn}, for the MTL setting of interest.
%
%
We formulate the training schedule as a Markov decision process which paves the way to employ policy learning methods 
to \emph{learn} the  scheduling policy.
We effectively and efficiently learn the training schedule policy within the imitation learning framework using   an  \emph{oracle policy} algorithm that dynamically sets the importance weights of auxiliary tasks based on their contributions to the generalisability of the main NMT task.
 Experiments on  low-resource  NMT settings  show the resulting automatically learned training schedulers are competitive with the best heuristics, and lead to up to +1.1 BLEU score improvements.
%

\end{abstract}

\section{Introduction}
Like many deep learning methods, NMT requires large amount of annotated data, i.e., bilingual sentence pairs, to train a model with a reasonable translation quality \cite{DBLP:journals/corr/KoehnK17}.
However, for many of the languages, we do not have the luxury of having large parallel datasets, a setting referred to as bilingually low-resource scenario.
Therefore, it is critical to compensate for lack of large bilingual train data using effective learning approaches.

Multi-task Learning (MTL) \cite{caruana1998multitask} is an effective approach to inject linguistic-based inductive biases into neural machine translation (NMT) in order to improve its generalisation and translation quality \cite{kiperwasser2018scheduled,DBLP:conf/acl/ZaremoodiBH18,zaremoodi2018neural}, specifically in bilingually low-resource scenarios.
%
%
Most of the recent literature has focused on the architectural design for effective MTL.
However, the beating heart of MTL i.e, \emph{training schedule} is less explored.
Training schedule is responsible to balance out the importance (participation rate) of different tasks throughout the training process, in order to make the best use of knowledge provided by the tasks.
Previous works on training schedule can be categorised based on the flavour of MTL they consider:
(1) \emph{general-MTL} where the goal is to improve all of the tasks \cite{chen2018gradnorm,guo2018dynamic};
(2) \emph{biased-MTL} where the aim is to improve one  of the tasks the most \cite{kiperwasser2018scheduled,zaremoodi2018neural}. 
Approaches in the latter category, including ours, try to devise the incorporation of auxiliary tasks, i.e., syntactic parsing, to improve the main task, even with the cost of degradation in their performance.

Recently, \cite{zaremoodi2018neural} proposed a \emph{hand-engineered} training schedule that considers a higher but \emph{fixed} participation rate for the translation task throughout the training.
\cite{kiperwasser2018scheduled} proposes a hand-engineered schedule to \emph{dynamically} change the participation ratio of the main task. 
Although, the results show improvements over the fixed training schedule on standard setting, their experimental setup is limited to only syntactic auxiliary tasks, and MTL with fully shared parameters of RNN-based architectures. 
More importantly, the schedule is restrictive as it needs to be (re-)designed for any learning scenario with different sets of auxiliary tasks.
To the best of our knowledge, there is no approach for automatically \emph{learning} a dynamic training schedule for biased-MTL.

In this paper, we propose a novel framework for automatically \emph{learning} how to multitask learn to maximally improve NMT as the main task.
This is achieved by formulating the problem as a Markov decision process (MDP), enabling to treat the training scheduler as the policy. 
We solve the MDP by proposing an effective yet computationally expensive \emph{oracle policy} that sets the participation rates of auxiliary tasks with respect to their contributions to the generalisation capability of the main translation task.
In order to scale up the decision making, we use the oracle policy as a teacher to train a scheduler network within the imitation learning framework using \mdag \cite{ross2011reduction}). 
%
%
Our experimental results on low-resource (English to Vietnamese/Turkish/Spanish) setting show up to +1 BLEU score improvements.
%
%

To summarise, our contributions are as follows:
\begin{itemize}
	\item We propose a novel framework for \emph{learning} the training schedule in MTL by formulating it as an MDP. 
	\item We propose an algorithmic \emph{oracle policy} that \emph{adaptively} and \emph{dynamically} selects tasks throughout the training. As the algorithmic oracle is computationally demanding, we scale up our approach by introducing a scheduler policy trained/used simultaneously using imitation learning to mimic the oracle policy. 
	\item 
	We evaluate our approach in low resource bilingual data scenarios using RNN architectures, and show up to +1 BLEU score improvements on three language pairs. 
\end{itemize}


\section{MTL as a Markov Decision Process}

\textbf{MTL Setup} 
Suppose we are given a set of a main task along with $K$ auxiliary tasks, 
each of which with its own training set $ \mathcal{D}^{k} :=  \{(\vx_i^{(k)},\vy_i^{(k)})\}_{i=0}^{N_{k}}$, where $k=0$ denotes the main task.  
We are interested to train a probabilistic MTL architecture $P_{\vTheta_{mtl}}(\vy|\vx)$, which accurately maps $\vx \in \mathcal{X}^{k}$ to its corresponding $\vy \in \mathcal{Y}^{k}$ for each task $k$. Without loss of generality, we assume that the model parameters $\vTheta$ are fully shared across all the tasks. 
The MTL parameters are then learned by maximising the log-likelihood  objective:
\begin{equation}
\label{eq:mtl-general}
    \arg\max_{\vTheta_{mtl}}\sum_{k=0}^{K}  w^{(k)} \sum_{i=0}^{N_k} \log P_{\vTheta}(\vy_i^{(k)}| \vx_i^{(k)}).
\end{equation} 
Typically, the tasks are assumed to have a predefined importance, e.g.  their weights are set to the uniform distribution  $w^{(k)} = \frac{1}{K+1}$, or set proportional to the size of the tasks' training datasets. 

Assuming an iterative learning algorithm, the standard steps in training the MTL architecture at time step $t$ are as follows: (i) A collection of training minibatches $\vb_t := \{\vb_t^{k}\}_{k=0}^{K}$ is provided, where $\vb_t^{k}$ comes from the $k$-th task, and (ii)  The MTL parameters  are re-trained by combining  information from the minibatches according to  tasks' importance weights. For example, the minibatch gradients can be  combined according to the tasks' weights, which can then be used to update the parameters using a gradient-based optimisation algorithm. 
\\
\\
\textbf{Markov Decision Process Formulation} There is evidence \cite{kiperwasser2018scheduled} that tasks' weights in the MTL training objective (eqn \ref{eq:mtl-general}) should be \emph{dynamically} changed during the training process to maximize the benefit of the main task from the auxiliary tasks. 
Intuitively, dynamically changing the  weights provides a mechanism to inject proper linguistic knowledge to the MTL architecture, in order to maximally increase the  generalisation capabilities on the main task. However, it is not clear how to set these weights to achieve this desiderata. \cite{kiperwasser2018scheduled} proposes \emph{hand-engineered policies} to change these weights according to predefined schedules.
%

\begin{figure}[t!]
\centering
  \includegraphics[scale=0.45]{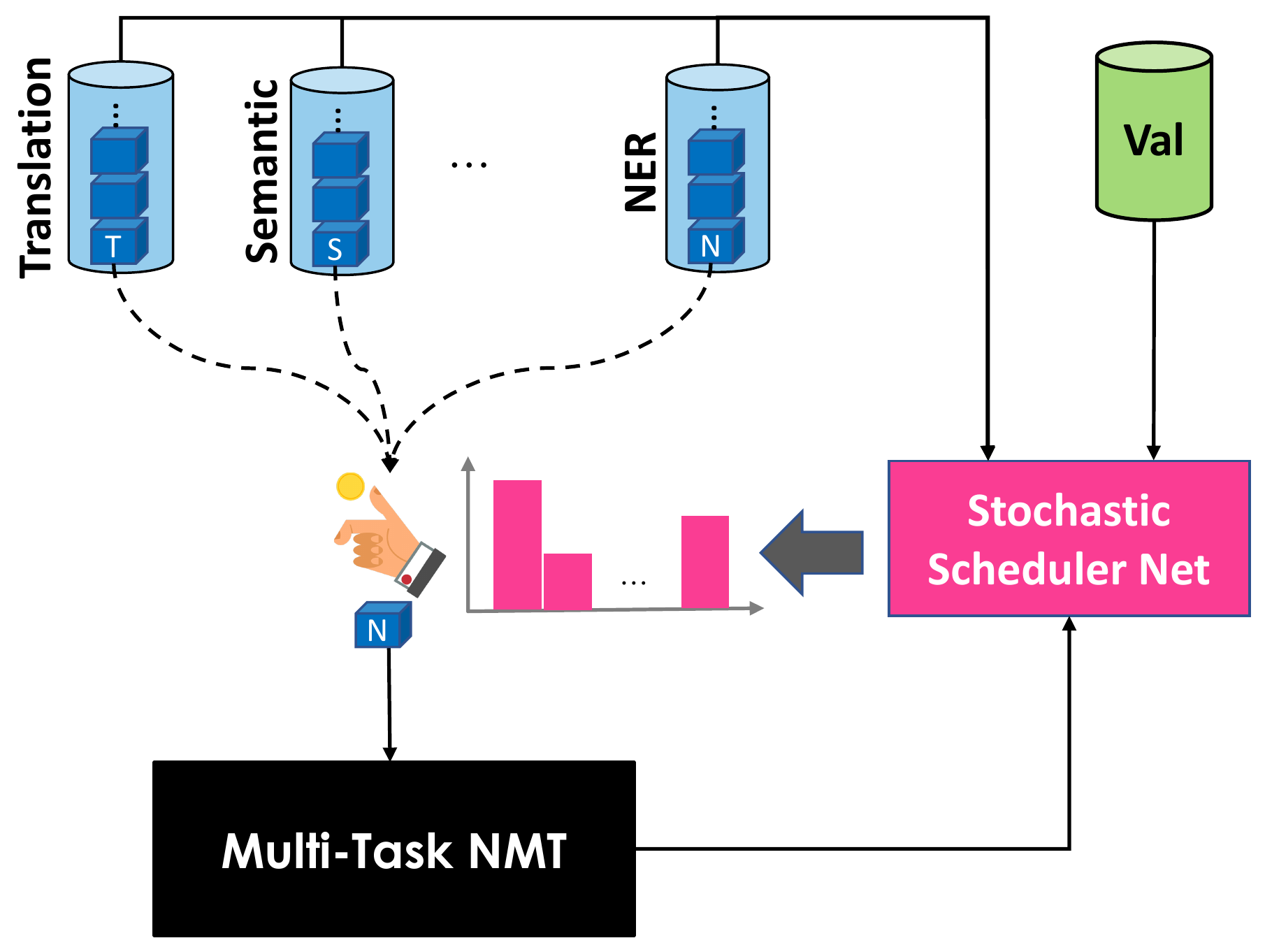}
  \caption{Overview of training an  MTL architecture using adaptive scheduling. Translation is the main task with syntactic and semantic parsing as auxiliary linguistic tasks.}
  \label{fig:meta-mtl}
\end{figure}

In this paper, we automatically learn MTL training  policies/schedules from  the data, in order to optimally change tasks' importance weights (see Figure \ref{fig:meta-mtl}). 
%
%
To achieve this goal, we formulate the training process of an MTL architecture as a Markov decision process (MDP), where its elements $(S,A,Pr(\vs_{t+1}|\vs_t,\va_t),R)$ are as follows. \\
$\bullet$ At the time step $t$ in the training process, the state $\vs_t$ includes any feature which summarises the history of the training trajectory. For example, it may include  the loss values encountered during training of the intermediate models, or any footprint of  the model parameters. The state space $S$ is accordingly specified by these features. \\
%
%
$\bullet$ Provided with a collection of training minibatches  $\vb_t$, 
the \emph{action} $\va_t$ then corresponds to the tasks' importance weights. That is,  the action space $A$ is the $K$-dimensional simplex $\triangle^{K}$, consisting of $K+1$ dimensional vectors with non-negative elements where the sum of the elements is one. \\
$\bullet$ $Pr(\vs_{t+1}|\vs_t,\va_t)$ is the transition function, which determines updated model parameters in the next time step, having  an action $\va_t \in A$ taken at the state $\vs_t \in S$. In other words, it corresponds to the update rule of the underlying optimisation algorithm, e.g. Stochastic Gradient Descent (SGD), having decided about the tasks' weights in the MTL training objective (eqn \ref{eq:mtl-general}) in the current state. \\
$\bullet$  $R(\vs_t,\va_t,\vs_{t+1})$ specifies the  instantaneous reward, having  an action $\va_t$ taken from the state $\vs_t$ and transitioned to the new state $\vs_{t+1}$. In our case, it should show the increase in the generalisation capabilities on the main task, having decided upon the tasks' importance weights, and accordingly updated the MTL model parameters. It is not trivial how to formalise and quantify the increase of the generalisation of the MTL architecture on the main task. Our  idea is to use a 
held-out validation set from the main task $\mathcal{D}^{val}$, and then use a loss function on this set to formalise the generalisation capability. More specifically, we take $-loss(\vTheta_{t+1},\mathcal{D}^{val})+loss(\vTheta_t,\mathcal{D}^{val})$ as a proxy for the increase in the generalisation capability of the main task.    \\
%
$\bullet$ Having set up the MDP formulation, our aim is to find the \emph{optimal} policy producing the best MTL training schedule $\va_t = \pi_{\vphi}(\vTheta_t,\vb_t)$, where $\vphi$ is the parameter of the \emph{policy network}. The policy prescribes what should be the importance of the MTL tasks at the current state, in order to get the best performing model on the main task in the long run.  
We consider $\pi_{\vphi}(\vTheta_t,\vb_t) \in \triangle^K$ as a probability distribution for selecting the next task and its training minibatch. It gives rise to a stochastic policy for task selection.

The optimal policy is found by maximising the following objective function: 
\vspace{-2.5mm}
\begin{equation}
    \arg\max_{\vphi} 
    \mathbb{E}_{\pi_{\vphi}}
    \Big[  
    \sum_{t=0}^{T} R(\vs_t,\va_t,\vs_{t+1}) 
    \Big]
\end{equation}
where $T$ corresponds to the maximum training steps for the MTL architecture.
%
Crucially, maximising the above 
long-term reward objective  corresponds 
to finding a policy, under which the validation loss of the resulting MTL model (at the end of the training trajectory) is minimised. 
In this paper, we will present methods to provide such optimal/reasonable stochastic policies for training MTL architectures.  
\section{An Oracle Policy for MTL-MDP}

In this section, we provide an \emph{oracle} policy ${\pi}^{\textrm{oracle}}$, which gives  an approximation of the optimal policy. 
The basic idea is to find an importance weight vector for the tasks which minimises the loss of the main task on a validation set.  
In other words, our oracle policy \emph{rolls-out} for one step from the current state, in order to reduce computational complexity.
However, we note that it is still extremly computationally demanding to compute oracle actions.
Therefore, we scale up our approach in the next section using a scheduler policy which mimics the oracle policy.

More specifically, our oracle learns the optimal tasks' weights $\vw_{opt}$ based on the following optimisation problem:

\begin{ChangeMargin}{-10.0em}{-22.0ex}%
{\small
\begin{empheq}[box=\fbox]{align}
& \vw_{opt} := \arg\min_{\hat{\pmb{w}} \in \triangle^K} 
\underbrace{-\sum_{\mathclap{(\vx,\vy) \in \vb^{val}}} 
\log P_{\hat{\vTheta}(\hat{\pmb{w}})}(\vy|\vx)}_{\mathcal{L}(\hat{\vTheta}(\hat{\vw}))}  \nonumber \\
\vspace{-1.5mm}
& \textrm{such that} \nonumber \\
 & \hat{\vTheta}(\hat{\pmb{w}}) :=  
 {\vTheta}_t +   \eta \sum_{k=0}^K \hat{w}^{(k)} \sum_{i=0}^{|\vb^{k}|-1} 
 \nabla \log P_{\vTheta_t }(\vy_i^{(k)}|\vx_i^{(k)}).  \nonumber
\end{empheq}
}
\end{ChangeMargin}
where we use a minibatch $\vb^{val}$ from the validation set for computational efficiency.
%
%
To find the optimal importance weights, we do one-step projected gradient descent on the objective function $\mathcal{L}(\vTheta_t)$ starting from zero. 
That is, we firstly set $\hat{\pmb{w}}_{opt} := \nabla_{\hat{\vw}_0}  \mathcal{L}(\hat{\vTheta}(\hat{\vw}_0)) \Bigr|_{\hat{\vw}_0=\pmb{0}}$, then zero out the negative elements of $\hat{\pmb{w}}$, and finally normalise the resulting vector to project back to the simplex and produce $\vw_{opt} \in \triangle^K$. 
Our approach to define $\hat{\vw}$ is based on the classic notion of \emph{influence} functions from robust statistics \cite{koh2017understanding,ren2018learning,cook1980characterizations}.

Computing $\nabla_{\hat{\vw}_0}  \mathcal{L}(\hat{\vTheta}(\hat{\vw}_0))$ is computationally expensive and complicated, as it involves backpropagation wrt $\hat{\vw}_0$ to a function which itself includes backpropagation wrt $\vTheta$. 
Interestingly, it can be proved that the component $\hat{w}^{(k)}_{opt}$ is proportional to the inner-product of the gradients of the loss over the minibatches from the validation set and the $k$-th task\footnote{See the supplementary for the proof.},

\vspace{-1.5mm}
{\small
$$ \sum_{j=0}^{|\vb^{val}|-1} 
 \nabla \log P_{\vTheta_t }(\vy_j^{(val)}|\vx_j^{(val)}) \cdot \sum_{i=0}^{|\vb^{k}|-1} 
 \nabla \log P_{\vTheta_t }(\vy_i^{(k)}|\vx_i^{(k)}).$$
 }
This approach is also computationally expensive and limits the scalability  of the oracle, as it requires backpropagation over the minibatches from \emph{all} of the tasks and validation set. 

%
%
%
\section{Learning to Multi-Task Learn}
In the previous section, we present an effective while computationally expensive oracle policy to solve the MTL-MDP.
In this section, we aim to scale up the decision making by \emph{learning} an efficient \emph{policy} to properly schedule the training process of the MTL architecture, i.e. learning to multi-task learn.  
Our MTL-MDP formulation paves the way to make use of a plethora of algorithms in imitation learning (IL) and reinforcement learning (RL) to learn the policy. 
It has been shown we can expect potentially \emph{exponentially} lower sample complexity in learning a task with IL than with RL algorithms \cite{sun2017deeply}.
Therefore, in this paper, we are going to explore the use of \mdag \cite{ross2011reduction}, a simple and effective algorithm for learning the policy within the IL framework.
In what follows, we first describe the architecture of our policy network, and then mention its training using \mdag.

%
%
%

\begin{figure}[t!]
\centering
  \includegraphics[scale=0.4]{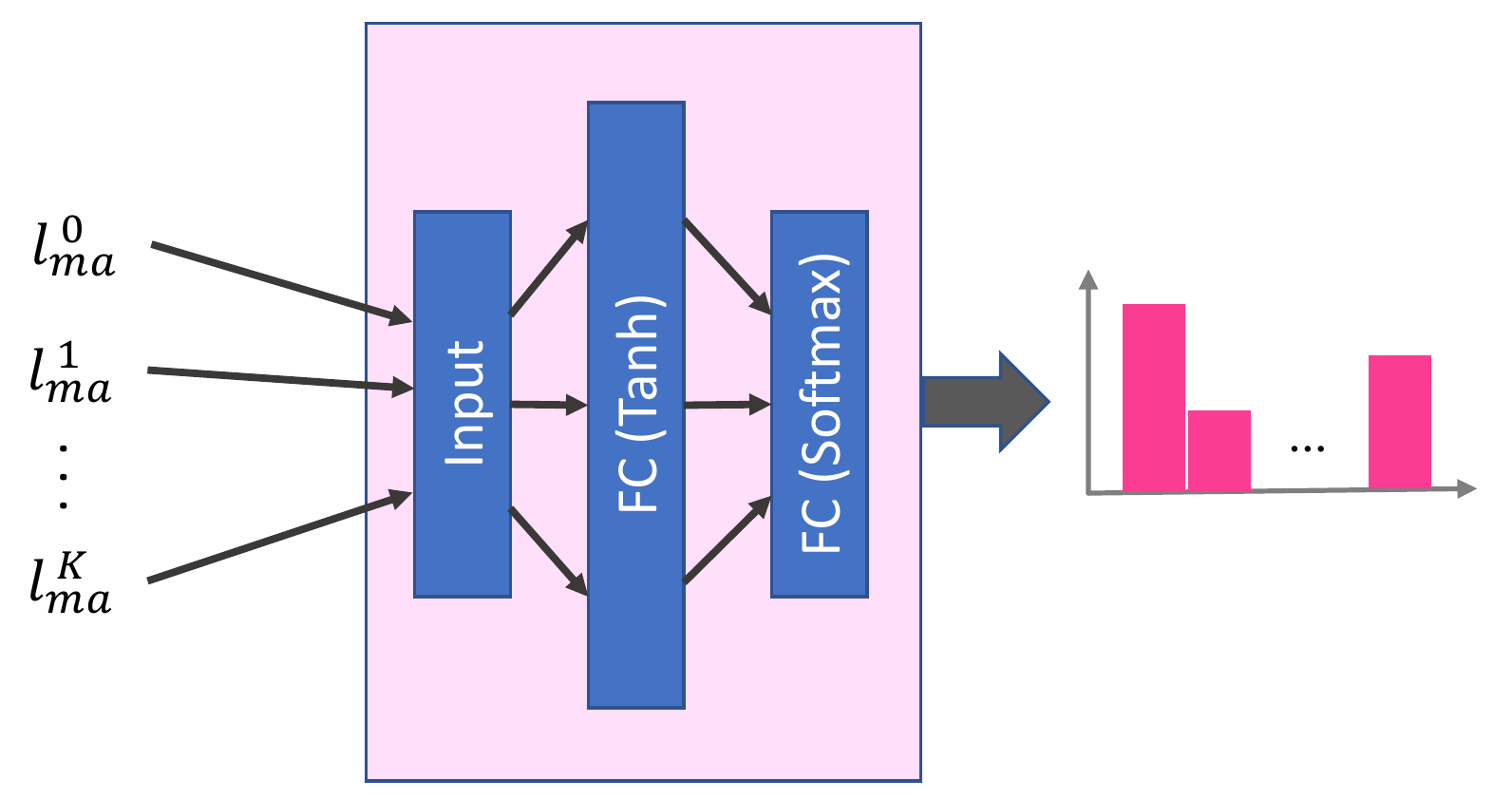}
  \caption{The policy/scheduler network.}
  \label{fig:schedule-net}
\end{figure}

\paragraph*{Policy/Scheduler Network}
We adopt a two-layer dense feed-forward neural network, followed by a Softmax layer, as the policy network (see Figure \ref{fig:schedule-net}). This is motivated by preliminary experiments on different architectures, which show that this architecture is effective  for the MTL scenarios in this paper. 

The inputs to the network include the footprint of history of the training process until the current time step $\vs_t$ as well as the those for the provided minibatches as the possible actions $\vb_t$. 
As the computational efficiency is critical when using the schedule network for large-scale MTL scenario, we opt to use light features to summarise the history of the training trajectory since the beginning. 
For each task $k$, we compute the moving average of its loss values $l_{ma}^k$  over those time steps where a minibatch from this task was selected for updating the MTL parameters. 
These features only depend on the minibatches used in the history of the training process, and do not depend on the provided minibatches at the current time step $\vb_t$. 
%

These features are updated in online manner during the MTL training process. More specifically, let us assume that a minibatch from a task $k_t$ is selected at the time step $t$ to retrain the MTL architecture. After updating the parameters of the underlying MTL architecture, we update the moving average loss \emph{only} for this task as follows: 
$$l_{ma}^{k_t} \leftarrow (1-\gamma) l_{ma}^{k_t} + \gamma \textrm{loss}(\vTheta_t,\vb_t^{k_t})$$  
where $\gamma \in [0,1]$.  Importantly, the loss value is already computed when updating the MTL parameters, so this feature does not impose additional  burden on the computational graph. 
These features are inspired by those used in MentorNet \cite{jiang2018mentornet}, and our investigations with the oracle policy of \S 3 on  finding informative features predictive of the selected tasks. 
%
%
%


\paragraph*{Learning the Policy with IL.}

Inspired by the Dataset Aggregation  (\mdag) algorithm \cite{ross2011reduction}, we learn the scheduler network jointly with the MTL architecture, in the course of a single training run.
Algorithm \ref{alg:high-IL} depicts the high-level procedure of the training and making use of the scheduler network.

At each update iteration, we decide between using or training the scheduler network with the probability of $\beta$\footnote{For efficiency, we \emph{use} the scheduler network at least 90\% of times.} (line 9).
In case of training the scheduler network (lines 12-14), we use the oracle policy $\pi^{\textrm{oracle}}$ in \S 3 to generate the optimal weights.
This creates a new training instance for the policy network, where the input is the current state and the output is the optimal weights. We add this new training instance to the memory replay buffer $M$, which is then used to re-train the policy/scheduler network.
In case of making use of the  scheduler network (line 10), we simply feed the state to the network and receive the predicted weights.

After getting the tasks' importance weights, the algorithm samples a task (according to the distribution $\vw_t$) to re-train the MTL architecture and update the moving average of the selected task (lines 16-18).  
\tiny{
\begin{algorithm}[t]
\caption{Learning the scheduler and MTL model}\label{alg:high-IL}
\begin{algorithmic}[1]
\State{\poorya{\tb{Input}: Train sets for the tasks $\mathcal{D}^{0}..\mathcal{D}^{K}$, validation of the main task $\mathcal{D}^{val}$, scheduler usage ratio $\beta$}}
\State{Init $\vTheta_0$ randomly} \Comment{MTL arch params}
\State{Init $\vphi$ randomly} \Comment{scheduler net params}
\State{$M \leftarrow \{\}$} \Comment{memory-replay buffer}
\State $l_{ma}^k \leftarrow 0 \ \ \ \ \forall k \in \{0,..,K\}$
\State $t \leftarrow 0$
\While{the stopping condition is not met}
\State {\small$\vb_t^0,..,\vb_t^k,\vb^{val} \leftarrow \textrm{sampleMB}(\mathcal{D}^{0},..,\mathcal{D}^{K}, \mathcal{D}^{val})$}
	\If{$\textrm{Rand}(0,1) < \beta$}  
		\Statex\hspace{\algorithmicindent}{$\triangleright$ Use the scheduler policy}
	    \State $\vw_t \leftarrow \pi_{\vphi}(l_{ma}^0,..,l_{ma}^K)$
	\Else
	    \Statex\hspace{\algorithmicindent}{$\triangleright$ Train the scheduler policy}
		\State $\vw_t \leftarrow \pi^{\textrm{oracle}}(\vb_t^0,..,\vb^{K},\vb^{val},\vTheta_t)$
		\State{$M \leftarrow M + \{ ((l_{ma}^0,..,l_{ma}^K),\vw_t) \}$}		\State{$\vphi \leftarrow \textrm{retrainScheduler}(\vphi,M)$}

	\EndIf
	\State{$k_t \leftarrow \textrm{sampleTask}(\vw_t)$}
	\State{$\vTheta_{t+1}, \textrm{loss} \leftarrow \text{retrainModel} (\vTheta_t, \vb_t^{k_t}$)}
	\State $l_{ma}^{k_t} \leftarrow (1-\gamma) l_{ma}^{k_t} + \gamma \textrm{loss}$
	\State $t \leftarrow t + 1$
\EndWhile

\end{algorithmic}
\end{algorithm}
}
\normalsize{
}

\paragraph*{Re-training the Scheduler Network} 
To train our policy/scheduler network, we optimise the parameters such
that the the action prescribed by
the network matches that which was prescribed by the oracle. 
More specifically, let $M := \{(\vs_i,\va_i)\}_{i=1}^I$ be the collected states paired with their optimal actions. The state $\vs_i$ comprises of  moving averages for the tasks $(l_{ma}^{i,0},..,l_{ma}^{i,K})$, and its paired  action is the optimal tasks' importance weights $\vw_i$. 
The training objective is 
$\min_{\vphi} \sum_{i=1}^I \textrm{loss}(\va_i, \pi_{\vphi}(\vs_i))$,  where we explore KL-divergence and $\ell_1$-norm of the difference of the two probability distributions as the loss function in the experiments of \S 5. 
To update the network parameters $\vphi$, we select a random minibatch from the memory replay $M$, and make one SGD step based on the gradient of the training objective.  


\section{Experiments}

\begin{table*}[t]
\begin{minipage}{\textwidth}
\centering
\begin{tabular}{lcccccccc}
\hline
\cline{1-9}
\multicolumn{1}{c}{ } & \multicolumn{2}{c}{En$\rightarrow$Vi} & \multicolumn{2}{c}{En$\rightarrow$Tr} & \multicolumn{4}{c}{En$\rightarrow$Es} \\
\cline{2-9}
\multicolumn{1}{c}{ } & \multicolumn{2}{c}{BLEU}  & \multicolumn{2}{c}{BLEU} & \multicolumn{2}{c}{BLEU} & \multicolumn{2}{c}{METEOR} \\
\multicolumn{1}{c}{}  &  Dev & Test & Dev & Test & Dev & Test & Dev & Test \\
\hline
 MT only & 22.83 & 24.15 & 8.55 & 8.5 & 14.49 & 13.44 & 31.3  & 31.1 \\

\hline
MTL with Heuristic Schedule \\
\ \ \  + Uniform & 23.10 & 24.81 & 9.14 & 8.94 & 12.81 & 12.12 & 29.6  & 29.5  \\
\ \ \  + Biased (Constant) & 23.42 & 25.22 & 10.06 & 9.53 & 15.14 & 14.11 & 31.8  & 31.3 \\

\ \ \  + Exponential & 23.45 & 25.65 & 9.62 & 9.12 & 12.25 & 11.62 & 28.0  & 28.1  \\
 \hline
 MTL with Adaptive Schedule \\
  \ \ \  + \textbf{SN} + heuristic & 23.86 & 25.70 & 10.53 & 10.18 & 13.20 & 12.38 & 29.9 & 29.7  \\
  \ \ \  + \textbf{SN} only & \tb{24.21} & \tb{26.45} & \tb{10.92} & \tb{10.62} & \tb{16.14} & \tb{15.12} & \tb{33.1} & \tb{32.7} \\

\hline
\end{tabular}
\caption{Results for three language pairs. ``+ SN'' indicates Scheduler Network is used in training. 
}
\label{tab:mtl-all}
\end{minipage}
\end{table*}

\subsection{Bilingual and Linguistic Tasks Corpora}

We analyse the effectiveness of our scheduling method on MTL models learned on languages with different underlying linguistic structures, and under different data availability regimes.
As the main focus of this paper is low-resource translation, we use three low-resource language-pairs English to Vietnamese/Turkish/Spanish\footnote{We use 150K training data for Spanish to simulate low-resource regime.}; see Table \ref{tab:mtl-data} for corpora details.
%
%
For a fair comparison, as the baselines do not have the scheduler, we add Val sets to their training sets.
For each of the language-pairs, we use BPE \cite{sennrich2015neural} with 40K operations. Using 
\textsf{fairseq}'s\footnote{https://github.com/pytorch/fairseq/blob/master/examples/translation} 
standard preparation pipeline, we filter out sentences longer than 250 tokens and sentence-pairs whose lengths ratio is more than 1.5. 


\begin{table}
\small
\centering
\begin{tabular}{l|c|c|c}
\hline
\cline{1-4}
 & En$\rightarrow$Vi & En$\rightarrow$Tr & En$\rightarrow$Es \\ 
\hline
Train & IWSLT 2015 & WMT 2016 & Europarl \\
Size & 133K & 200K & first 150K  \\
Dev & tst2012 & NT2016 & NT2011 \\
Val & tst2012 & NT2017 & NT2012 \\
Test & tst2013 & NT2018 & NT2013\\
\hline
\end{tabular}
\caption{Details of bilingual corpora. ''NT'' stands for ''newstest''. Dev is used for early stopping, and Val is used by oracle policy for training the scheduler network.}
\label{tab:mtl-data}
\end{table}




%

\paragraph*{Auxiliary Linguistic Tasks}
In order to inject syntactic and semantic knowledge into NMT we incorporate the following linguistic tasks:
(1.) Named-Entity Recognition (NER): 
We use CoNLL shared task data\footnote{https://www.clips.uantwerpen.be/conll2003/ner} which  concentrates on four types of named entities: persons, locations, organisations and names of miscellaneous entities.
(2.) Syntactic Parsing: We use Penn Treebank parsing data with the standard split for training, development, and test \cite{Marcus:1993}.
As the parsed sentences are in the tree format, we have linearised them \cite{NIPS2015_5635}.
(3.) Semantic parsing: We use the abstract meaning representation (AMR) corpus Release 2.0 (LDC2017T10)\footnote{https://catalog.ldc.upenn.edu/LDC2017T10}.
We have applied BPE with 40K operations on the linguistic corpora and followed the same preprocessing procedure used for bilingual corpora.




\subsection{MTL architectures}
As recent studies on the sharing strategy for MTL models have shown that the partial sharing is more effective than full sharing \cite{Liu2017Adversarial,Guo2018Soft,zaremoodi2018neural}, we share a partial fraction of model parameters in our experiments.
We have implemented our method and baselines using PyTorch \cite{pytorch} on top of OpenNMT \cite{opennmt}. 
For the scheduler network, the number of hidden dimensions and the decay factor ($\gamma$) are set to 200 and 0.7, respectively.
For its training, we use L1 loss function, and Adam optimiser with learning of 0.0001.
For low-resource language pairs we use LSTM setting while for the high resource scenario we have used Transformer.
The $\beta$ (ratio of using over training the scheduler) is set to 0.99/0.9 for Transformer/LSTM settings.
%

 We use the architecture proposed in \cite{zaremoodi2018neural} with 3 stacked LSTM layers with 512 hidden nodes in encoders and decoders.
The batch size and dropout are set to 32 and 0.3, respectively.
For sharing, we have shared 1 top layer of encoders, 2 bottom layers of decoders, and vocabularies among the tasks.
For optimisation, we use Adam optimiser with 0.001 as its learning rate.
All models are trained for 25 epochs on a single NVIDIA V100 GPU and the best models are saved based on the perplexity on the Dev set.

\paragraph*{\emph{Hand-engineered} training schedules.}
At each update iteration, training schedules determine the probability distribution over tasks for selecting the source of the next training mini-batch.
Considering a slope parameter $\alpha$ and the fraction of training epochs done so far $t=sents/||corpus||$, the schedulers are as follows:
\begin{itemize}
    \item \tb{Uniform}: Selects a random minibatch from each task.
    \item \tb{Biased} \cite{zaremoodi2018neural}: Selects two random minibatches, one from the main task (translation) and the other from a randomly selected task. Thus, the weight of the main task would be as same as the sum of auxiliary tasks weights. This is similar to Constant scheduler in \cite{kiperwasser2018scheduled},$P_{m}(t)=\alpha=0.5$\footnote{Similar to their experiments, we have set $\alpha$ to 0.5.}, where $P_{m}(t)$ is the probability of selection of main task.
    \item \tb{Exponential:} In this schedule the probability of selecting the main task starts from 0 and exponentially grows i.e. $P_{m}(t)=1-e^{-\alpha t}$. The rest of the probability will be divided uniformly among the auxiliary tasks.
    %
\end{itemize}

\reza{
\paragraph*{Mixture of hand-engineered and the proposed \emph{automatic} training schedules.}
To the best of our knowledge, there is no hand-engineered MTL training schedule for NMT which distinguishes among auxiliary tasks.
These schedules only balance out the participation of the main translation task versus auxiliary tasks, and assign uniform participation probability among auxiliary tasks.
Hence, we created a new schedule by combining a hand-engineered schedule along with the proposed policy/scheduler network to see the effectiveness of their combination.
In this schedule, the probability of selection of the main task is determined by a hand-engineered schedule (Biased for LSTM setting and Exponential for Transformer setting).
However, instead of uniformly distributing the remaining probability among the auxiliary tasks, we use the scheduler network to assign a probability to \emph{each} of the auxiliary tasks based on their contribution to the generalisation of the MTL model.
}

\subsection{Results}
Table \ref{tab:mtl-all} reports the results for  our proposed method and the baselines for the bilingually low-resource conditions, i.e. translation from English into Vietnamese/Turkish/Spanish.
As seen, the NMT models trained with our scheduler network perform the best across different language pairs.
More specifically, the three MTL training heuristics are  effective in producing models which outperform the MT-only baseline. 
Among the three heuristics, the Biased training strategy is more effective than Uniform and Exponential, and leads  to trained models with substantially better translation quality than others.  
Although our policy learning is agnostic to this MTL setup, it has \emph{automatically} learned  effective training strategies,   leading to further improvements compared to Uniform as the best  heuristic training strategy. 
We further considered learning a training strategy which is a combination of the best heuristic (ie Biased) and the scheduler network, as described before. 
As seen, this combined policy is not as effective as the pure scheduler network, although it is still better than the best heuristic training strategy.  
%


\section{Related Work}

Multi-task learning has attracted attention to improve NMT in recent work. %
\cite{DBLP:conf/acl/ZaremoodiBH18,zaremoodi2018neural} have explored the use of syntactic parsing, semantic parsing, and NER to improve the performance of NMT in low-resource scenarios.
\cite{niehues2017exploiting} has made use of part-of-speech
tagging and named-entity recognition tasks to
improve NMT. 
%
%
More broadly, MTL has been used for various NLP problems, e.g.  dependency parsing \cite{peng-etal-2017-deep}, video captioning \cite{pasunuru2017multitask}  
key-phrase boundary classification \cite{DBLP:conf/acl/AugensteinS17}, 
Chinese word segmentation, and  text classification problem \cite{Liu2017Adversarial}.

A crucial problem in MTL is the negative transfer phenomena. 
One line of research aims to learn effective  MTL architectures  for sharing the parameters, e.g. \cite{DBLP:conf/acl/ZaremoodiBH18,DBLP:journals/corr/RuderBAS17}.
Another approach is to learn effective MTL training schedules, which includes our approach.
\cite{kiperwasser2018scheduled} has proposed the use of hand-engineered training schedules for MTL in NMT. 
%
%
\poorya{
\cite{DBLP:journals/corr/abs-1904-04153} has proposed an approach to automatically determine weights of tasks in the MTL training.
Weights are tuned via a guided grid search where the performance for each set of weights is modelled as a sample from a Gaussian Process (GP).
It needs be re-trained (from scratch) the MTL model for each set of weights which is computationally exhaustive.
Moreover, the weights are \emph{fixed} inside each run.
In contrast, our approach is based on \emph{adaptive} weights during the MTL training where the weights are trained/used in the course of only a \emph{single} run.
\cite{NIPS2018_7819} 
has proposed a learning-to-MTL  framework in order to  learn effective MTL \emph{architectures} for generalising to  new tasks. 
This is achieved by collecting historical multitask experience, represented by tuples consisting of the MTL problem, MTL architecture, and its relative error. 
In contrast, our learning-to-MTL framework tackles the problem of learning effective training schedules to use auxiliary tasks in such a way to improve the main translation task the most. 
}
%


{
\section{Conclusions}
We  introduce a novel approach for automatically learning effective training schedules for MTL.  
We formulate MTL training as a Markov decision process, paving the way to treat the training scheduler as the policy.
We then introduce an effective \emph{oracle policy}, and use it to train a policy/scheduler network within the imitation learning framework using \mdag in on-policy manner.
%
%
Our  results on low-resource (English to Vietnamese/Turkish/Spanish) settings using LSTM  architectures show up to +1.1 BLEU score improvement.
%
%
Although the focus of the paper is on NMT, the presented approach is general and can be applied to any MTL problem, which we leave as the future work.
}

\bibliography{ref.bib}
\bibliographystyle{aaai}

\end{document}